\documentclass{article}

 \PassOptionsToPackage{numbers}{natbib}
\usepackage[final]{neurips_2023}

\usepackage[utf8]{inputenc} 
\usepackage[T1]{fontenc}    
\usepackage{hyperref}       
\usepackage{url}            
\usepackage{booktabs}       
\usepackage{amsfonts}       
\usepackage{nicefrac}       
\usepackage{microtype}      
\usepackage{xcolor}         
\usepackage{booktabs}
\usepackage{graphicx}
\usepackage[numbers]{natbib}

\title{DUPE: Detection Undermining via Prompt Engineering for Deepfake Text}

\author{
  James Weichert\\
  Virginia Tech\\
  \texttt{jamesweichert@vt.edu}\\
  \And
  Chinecherem Dimobi\\
  Virginia Tech\\
  \texttt{chinedimobi@vt.edu@vt.edu} \\
  }

\begin{document}

\maketitle

\begin{abstract}
  As large language models (LLMs) become increasingly commonplace, concern about distinguishing between human and AI text increases as well. The growing power of these models is of particular concern to teachers, who may worry that students will use LLMs to write school assignments. Facing a technology with which they are unfamiliar, teachers may turn to publicly-available AI text detectors. Yet the accuracy of many of these detectors has not been thoroughly verified, posing potential harm to students who are falsely accused of academic dishonesty. In this paper, we evaluate three different AI text detectors—Kirchenbauer et al. watermarks, ZeroGPT, and GPTZero—against  human and AI-generated essays. We find that watermarking results in a high false positive rate, and that ZeroGPT has both high false positive and false negative rates. Further, we are able to significantly increase the false negative rate of all  detectors by using ChatGPT 3.5 to paraphrase the original AI-generated texts, thereby effectively bypassing the detectors. 
\end{abstract}

\section{Introduction}

The task of deepfake text detection focuses on distinguishing text written by humans from that created by machines, especially Large Language Models (LLM) like OpenAI's GPT model family. These foundation models are designed to process and respond to a variety of input prompts, producing high-quality text across a diverse range of topics. LLM tools are becoming widely used, particularly by students looking for inspiration on an essay topic or looking to have an LLM write their essay entirely. The potential for misuse of these LLMs raises concerns for educators about academic integrity and originality of a student's work. Previous research has shown that LLMs can be used to create fake news articles, social media posts, or emails that appear to be written by real people or organizations \cite{pu_deepfake_2023}. As we show in this paper, LLMs are also capable of structuring if not fully completing college-level academic essays. However, the reliability of current AI text detection tools is questionable, and we note that the potential harm from false accusations of academic dishonesty are severe and likely to be unevenly distributed. For example, Liang et al. \cite{liang_gpt_2023} show that GPT text detectors are biased against non-native English writers.

Several methods have been proposed to detect AI text, including those that rely on statistical, linguistic, and contextual features, like DetectGPT \cite{mitchell_detectgpt_2023}. However, statistical methods often mistakenly flag genuine text as fake \cite{pu_deepfake_2023}. Likewise, machine learning-based approaches may struggle to adapt to different text domains \cite{chakraborty_possibilities_2023, wang_generalizing_2021}. These limitations suggest that current methods might not be fully effective in practical situations. Presently, newer strategies like watermarking are emerging \cite{kirchenbauer_watermark_2023, kirchenbauer_reliability_2023}. 

In this paper, we examine three different detection methods—Kirchenbauer et al. watermarking, ZeroGPT, and GPTZero—using human and AI-written essays. We show that these detectors fall short of their claims about accuracy and error rates. In short, our contributions in this paper are twofold: (1) we synthesize a dataset of 212 human-written and 208 ChatGPT-generated academic essays across four college disciplines; and (2) we successfully deceive three common AI text detectors into misclassifying AI-generated text as human-written using only ChatGPT paraphrasing.

\subsection{Prior work}

Previous studies have explored the reliability of AI-generated text detectors. One such study by Sadasivan et al.\cite{sadasivan_can_2023} used PEGASUS, a streamlined neural network-based paraphrasing tool, on 100 AI-generated, watermarked texts. The authors showed significant results after five rounds of recursive paraphrasing. Another study by Weber-Wulff et al. \cite{weber-wulff_testing_2023} evaluated 14 tools—including ZeroGPT and GPTZero—for their ability to detect text generated by ChatGPT. The researchers tested both human-written and AI-generated texts, as well as those edited by humans and AI. Their findings revealed that none of the detectors reached 80\% accuracy, with only five surpassing 70\%.

Kirchenbauer et al., in their second paper on watermarking \cite{kirchenbauer_reliability_2023}, investigated the resilience of watermarked texts under various modifications. They tested watermarked texts that were either rewritten by humans, paraphrased by an LLM, or blended into longer, manually written documents. The authors make strong claims that watermarks remain detectable despite these changes. They note that even though modifications may weaken watermarks, paraphrased texts often keep enough original content to enable watermark detection, particularly when examining a large portion of the text.

While previous work has examined the accuracy of the AI text detectors we use in this paper, we extend on prior work by focusing on the prompt engineering required to induce ChatGPT to paraphrase AI-generated texts to avoid detection. Additionally, we focus on the undergraduate academic domain by using 212 student-written papers and GPT essay generations based off of the original human texts. To our knowledge, we are the first to take this approach to research the reliability of AI text detectors.

\section{Deepfake text detection schemes}

\subsection{Watermarking} \label{watermarking}

Text watermarking is an emergent classification paradigm that involves altering text at generation time to embed a signature through vocabulary or sentence structure, which can be detected from the text at a later time \cite{sadasivan_can_2023}. The watermarking scheme described by Kirchenbauer et al. \cite{kirchenbauer_watermark_2023, kirchenbauer_reliability_2023} involves partitioning the LLM’s vocabulary into red and green lists according to the green list proportion, $\gamma$, then preferentially sampling tokens from the green list subject to semantic constraints. The authors claim that watermarking is robust against both machine and ``expert-level'' human paraphrasing.

We hypothesize that the Kirchenbauer watermarking scheme is not as robust as the authors claim, since it assumes that the distribution of words (tokens) in human text is effectively uniformly random, with the mean proportion of green listed tokens in human text equal to $\gamma$. This motivates the use of a p-value based on the distribution of means of green listed tokens, which the Central Limit Theorem guarantees will be normally distributed and centered at $\gamma$. A small p-value, then, is evidence of an abnormally high proportion of green listed tokens, and therefore is unlikely to be human-generated.

We argue that this assumption does not hold, primarily because human text creation is decidedly \textit{not} random. The proportion of green listed tokens in a human text is not roughly equal to $\gamma$, because words have different use frequencies; certain words are used much more often than others. While the best approximation of the distribution of word frequencies is still a matter of debate among linguists \cite{baayen_statistical_1992}, there is no contention that word frequencies are uniformly distributed, which is the only distribution under which Kirchenbauer et al.’s scheme should be expected to work. 

\subsection{Post-Hoc detectors}

While watermarking requires whitebox access to the LLM at generation time, post-hoc detectors rely only on a trained or human-configured understanding of how LLMs generate text in order to detect similar characteristics in unknown text samples \cite{kirchenbauer_watermark_2023}. However, the configuration of post-hoc detectors may not be disclosed when the detector is a paid product. For this reason, we evaluate the detection scores of two popular online AI text detectors—ZeroGPT and GPTZero—across a variety of texts in order to formulate a general hypothesis about how these classifiers function. Possibly the simplest form of a post-hoc detector is one that primarily uses a text’s linguistic perplexity for classification, assuming that LLMs generate text with lower perplexity than human text. Although the exact nature of its classifier is not disclosed, we suspect that ZeroGPT's detector model relies heavily on a text's perplexity. Figure \ref{fig:perplexity-scatter} in Appendix \ref{detector-descriptions} shows the relatively strong negative correlation between the text's perplexity and its ZeroGPT ``AI \%'' score.

\section{Methodology}

\subsection{Data} \label{data}

To evaluate the chosen AI text detection schemes on human text, we select a subset of human-written essays from the Michigan Corpus of Upper-Level Student Papers (MICUSP) \cite{the_regents_of_the_university_of_michigan_michigan_2009}. MICUSP consists of academic papers written by senior undergraduate and graduate students from a variety of University of Michigan courses. All essays were written between 2004 and 2009, ruling out the possibility that the texts were written with the help of LLMs. Our subset (n = 212) includes all undergraduate papers in the disciplines of English (n = 81), Biology (n = 47), Political Science (n = 51), and Philosophy (n = 33). These disciplines were chosen because of the large number of papers available for these subjects, and together the subset provides significant variation in the subject, style, and writing register.

In order to generate AI-written texts with similar subject matter and writing style to the human essays, we prompt the LLM with a standardized prompt created by using the discipline and title of the corresponding human essay in our MICUSP subset. For example, the AI text corresponding to the human English essay titled “The Vicar of Wakefield as a Failed Morality Story” was generated using the prompt: “Write a College English class essay titled ‘The Vicar of Wakefield as a Failed Morality Story’”. Using this standardized prompt, we can generate AI-written essays in a one-to-one ratio with the human essays, replicating the same diversity of writing styles, subjects, and vocabularies.


For watermarking, we used GPT Neo (1.3 billion parameters), since watermarking reqyures whitebox access to the generating model. We aligned our model's settings with those used in Kirchenbauer et al.\cite{kirchenbauer_watermark_2023}. For generating the watermarked essays, we used the first 10 words from each human essay in our collection as prompts. We also set a limit on the length of the generated texts at 1000 tokens.

\subsection{Text classification}

To test our detection evasion attacks, we choose three AI text detectors which we believe are representative of the current deepfake detection landscape: Kirchenbauer et al. \textit{Watermarking}, \textit{ZeroGPT\footnote{\href{https://www.zerogpt.com/}{https://www.zerogpt.com/}}}, and \textit{GPTZero\footnote{\href{https://gptzero.me/}{https://gptzero.me/}}}. With the exception of the Kirchenbauer et al. watermark scheme (which was accepted at ICML 2023), we do not necessarily argue that these detectors represent state-of-the-art deepfake text detection technology. However, we include ZeroGPT and GPTZero because they are two widely-available and widely-used detectors currently on the market; they are, at the time of writing, the second and first results on Google for the search ``chat gpt text detector,'' respectively. Thus, in the context of this research, these two detectors are useful in assessing the potential for misclassification when, for example, an instructor uses one such detector to check a student's essay for AI content. Appendix \ref{detector-descriptions} provides further detail on each of the three detectors used in this paper.

\subsection{Paraphrasing} \label{paraphrasing-method}

In order to deceive the detectors, we paraphrase the original AI-generated essays using ChatGPT 3.5. This simulates the process of a student using ChatGPT to paraphrase the essay before turning it in. For each AI-generated essay, we generate AI paraphrases using the following four prompts:\footnote{For each paraphrasing attack, ``[TEXT]'' is replaced with the original AI-generated text being paraphrased.}

\begin{enumerate}
    \item \textit{Perplexity} Paraphrasing: ``Paraphrase the following text to increase the average sentence length and the sentence perplexity: [TEXT]''
    \item \textit{Word Replacement} Paraphrasing: ``Rewrite the following passage, preserving the original meaning but using different words and sentence structures while keeping the same length of the original passage: [TEXT]''
    \item \textit{College Student} Paraphrasing: ``Rewrite the following text to make it sound like it was written by a college student. You can modify sentences, replace words, and make any editorial changes necessary to make the text more readable and simple:  [TEXT]''
    \item \textit{Recursive} Paraphrasing: ``Paraphrase the following text to make it easier to read and more human-sounding, while maintaining a formal writing register and the content of the original text:  [PERPLEXITY PARAPHRASED TEXT]''
\end{enumerate}

The \textit{Perplexity}, \textit{College Student}, and \textit{Recursive} paraphrasing attacks were each tested against ZeroGPT and GPTZero, while the \textit{Word Replacement} attack was only used for watermarking. We constructed these prompts heuristically, attempting to target aspects of the underlying text which we suspected were key to a detector's classifications. For watermarking, for example, we constructed prompt 2 to replace green listed words with synonyms, thereby diluting the green list proportion in the paraphrased text. For the other prompts, we hypothesize that the text's perplexity is a key determinant for classification, and thus we developed prompts 1, 3 and 4 as different ways of achieving the desired high perplexity text. Appendix \ref{text-examples} includes examples of the human essay, the original ChatGPT generation, and the \textit{Perplexity}, \textit{College Student}, and \textit{Recursive} paraphrases of the same text.

\subsection{Semantic robustness}

In order to measure the extent to which the paraphrased texts maintain roughly the same semantic content as the original AI-generated texts, we use Google's Universal Sentence Encoder (USE) developed by Cer et al \cite{cer_universal_2018} to measure semantic similarity. The Universal Sentence Encoder generates a standardized 512-dimensional embedding vector for each input, which can be multiplied with another embedding vector to compute a similarity score between the two texts. Since components of each vector sum to 1, this dot product is equivalent to the \textit{cosine similarity} metric. The Universal Sentence Encoder is useful because the it performs well on the Semantic Textual Similarity (STS) benchmark, which is correlated with human judgements about the similarity of two texts. A high cosine similarity between two texts should indicate that the texts are semantically similar.

\section{Experiments}

\subsection{Baselines: false positive and false negative rates}

First, we explore how the deepfake detectors perform on unaltered inputs, both human-written and LLM-generated. We calculate the false positive and false negative rates for each detector as follows:

\begin{enumerate}
    \item For Kirchenbauer watermarking, a \textit{positive} classification is defined as any p-value below a 5\% cutoff level. We also report the average p-value across all texts.
    \item For ZeroGPT, we define a \textit{negative} classification as a text receiving either the classification of “Your Text is Human written” or “Your Text is Most Likely Human written”. All other ZeroGPT classifications are considered positive classifications. We also report the “GPT \%” scores given by ZeroGPT.
    \item For GPTZero, we define a \textit{negative} classification as a text receiving an “AI Probability” score of less than 50\% (i.e. GPTZero concludes the text is more likely to be human than AI). We also report the “AI Probability” scores given by GPTZero.
\end{enumerate}

The false positive rate (FPR) across all essay disciplines for each of the three text detectors is shown in Table \ref{fpr-table}. While GPTZero has a 0\% FPR, the other two detectors perform quite poorly, with at least one in every five human texts being misclassified as an AI-generated text. The high false positive rate for watermarking in particular supports our hypothesis that the Kirchenbauer et al. watermarking scheme does not properly account for the distribution of word frequencies in human language. The false negative rate (FNR) across all essay disciplines is shown in Table \ref{fnr-table}. GPTZero again performs well here, as does watermarking. The nearly 20\% false negative rate on unaltered ChatGPT text combined with a near-25\% false positive rate already indicates that ZeroGPT is not a reliable classifier.

\begin{table}
  \caption{Detector false positive rates}
  \label{fpr-table}
  \centering
  \small
\begin{tabular}{lcccccc}\toprule

    & \multicolumn{2}{c}{\textbf{Watermarking}} & \multicolumn{2}{c}{\textbf{ZeroGPT}} & \multicolumn{2}{c}{\textbf{GPTZero}}

    \\\cmidrule(lr){2-3}\cmidrule(lr){4-5} \cmidrule(lr){6-7}
    Discipline & Avg. p-value & FPR & Avg. ``GPT \%'' & FPR & Avg. ``AI Prob.'' & FPR \\\midrule

    English & 0.471 & 16.04\% & 5.27\% & 3.70\% & 3.40\% & 0.00\% \\
    Biology & 0.485 & 19.14\% & 18.59\% & 31.91\% & 2.85\% & 0.00\% \\
    Political Science & 0.331 & 27.45\% & 16.10\% & 23.53\% & 2.25\% & 0.00\% \\
    Philosophy & 0.414 & 21.21\% & 20.26\% & 39.39 \% & 2.85\% & 0.00 \% \\\midrule

    Overall (Avg.) & 0.425 & \textbf{20.96\%} & 15.05\% & \textbf{24.64\%} & 2.84\% & \textbf{0.00\%} \\\bottomrule

\end{tabular}
\end{table}

\begin{table}
  \caption{Detector false negative rates}
  \label{fnr-table}
  \centering
  \small
\begin{tabular}{lcccccc}\toprule

    & \multicolumn{2}{c}{\textbf{Watermarking}} & \multicolumn{2}{c}{\textbf{ZeroGPT}} & \multicolumn{2}{c}{\textbf{GPTZero}}

    \\\cmidrule(lr){2-3}\cmidrule(lr){4-5} \cmidrule(lr){6-7}
    Discipline & Avg. p-value & FNR & Avg. ``GPT \%'' & FNR & Avg. ``AI Prob.'' & FNR \\\midrule

    English & 2.49e-23 & 0.00\% & 36.71\% & 36.25\% & 53.30\% & 0.00\% \\
    Biology & 1.67e-17 & 0.00\% & 49.98\% & 8.70\% & 59.65\% & 0.00\% \\
    Political Science & 2.21e-37 & 0.00\% & 57.16\% & 16.00\% & 71.20\% & 0.00\% \\
    Philosophy & 8.12e-21 & 0.00\% & 36.74\% & 15.63\% & 58.80\% & 10.00\% \\\midrule

    Overall (Avg.) & 4.18e-18 & \textbf{0.00\%} & 44.57\% & \textbf{19.14\%} & 60.73\% & \textbf{2.50\%} \\\bottomrule

\end{tabular}
\end{table}

\subsection{ChatGPT paraphrasing}

\begin{table}
  \caption{Attack success rates (FNR) for each attack}
  \label{asr-table}
  \centering
  \small
\begin{tabular}{lcccccc}\toprule

    & \multicolumn{2}{c}{\textbf{Watermarking}} & \multicolumn{2}{c}{\textbf{ZeroGPT}} & \multicolumn{2}{c}{\textbf{GPTZero}}

    \\\cmidrule(lr){2-3}\cmidrule(lr){4-5} \cmidrule(lr){6-7}
    Paraphrasing Attack & ASR (FNR) & FNR $\Delta$ & ASR (FNR) & FNR $\Delta$  & ASR (FNR) & FNR $\Delta$  \\\midrule

    \textit{Perplexity} & - & - & \textbf{89.64\%} & +70.50\% & 22.50\% & +20.00\% \\
    \textit{Word Replacement} & \textbf{91.55\%} & +91.55\% & - & - & - & - \\
    \textit{College Student} & - & - & 88.75\% & +69.61\% & \textbf{51.25\%} & +48.75\% \\
    \textit{Recursive} & - & - & 75.00\% & +55.86\% & 1.25\% & -1.25\% \\\bottomrule

\end{tabular}
\end{table}

We paraphrased the original LLM-generated essays using ChatGPT 3.5 according to the methodology described in section \ref{paraphrasing-method}. We define the attack success rate (ASR) as the false negative rate of the paraphrased texts, and report this rate along with the change in false negative rate (\textit{FNR $\Delta$}) compared to the baseline FNR in Table \ref{asr-table}. While different attacks perform better on different detectors, we were able to achieve a minimum 50\% attack success rate for all three detectors. The attack success rates for watermarking and ZeroGPT are particularly high, raising concerns about their reliability.

\subsection{Semantic similarity}

Table \ref{use-table} in Appendix \ref{semantic-similarity} summarizes the Universal Sentence Encoder cosine similarity scores for various combinations of text types. On average, any two human-written texts from the MICUSP corpus had a cosine similarity of 0.28. Using this as a benchmark, we see evidence in the other similarity scores that paraphrasing is, on the whole, maintaining the semantic content of the original text. Surprisingly, the average GPT text generation and its corresponding human essay has a  cosine similarity of 0.63, even though the content of the human essay was not included in the generation prompt. \textit{Perplexity} and \textit{Recursive} paraphrasing both have high cosine similarity with the original AI texts, while the \textit{College Student} paraphrasing has a slightly lower similarity value. We view this lower score as quantitative support for the observation that the \textit{College Student}-paraphrased texts use informal, colloquial language that deviates from the more formal register used in the original AI text.

\section{Discussion}

\subsection{AI text detectors: claims vs. reality}\label{detector-claims}

ZeroGPT performed the poorest, with an 89.64\% overall attack success rate using the best attack (\textit{Perplexity} paraphrasing), and 24.64\% and 19.14\% false positive and false negative rates. This results in an overall accuracy of 78.62\%, which does not match the ``up to 98\%'' accuracy rate claimed by the ZeroGPT website, leading us to seriously question the efficacy of the detector.

GPTZero fared much better, despite our 51.25\% attack success rate using the \textit{College Student} paraphrasing attack. GPTZero claims false positive and false negative rates of 4\%, which is higher than our findings of 0.00\% FPR and 2.50\% FNR. Additionally, the only AUC score of 0.98 mentioned on the GPTZero website was in line with the 1.0 AUC score we calculated using our results.

Kirchenbauer et al. watermarking fared much more poorly in our experiments than the claims in \cite{kirchenbauer_watermark_2023} and \cite{kirchenbauer_reliability_2023} would lead one to believe. While we found a false negative rate of 0\%, the 11.73\% false positive rate we calculated is many orders of magnitude greater than the 1e-5 rate to which Kirchenbauer et al. claim to be able to limit the watermarking scheme. The 11.73\% FPR supports our hypothesis that the watermarking algorithm does not accurately model human language by assuming that tokens in human text are uniformly distributed. This, coupled with our 92\% attack success rate using the \textit{Word Replacement} paraphrasing attack, indicates that further examination of the reliability of Kirchenbauer watermarking is warranted. Without knowing more about the specific text and paraphrasing examples used by the authors in \cite{kirchenbauer_reliability_2023}, it is difficult to assess why our results differ so much from those in the Kirchenbauer papers.

\subsection{Limitations} \label{limitations}

Our experiments represent realistic `real-world` attacks that could be employed, e.g., by a student looking to use ChatGPT to write an academic paper while avoiding detection. The tools we used are all available for free online and our attacks assumed no whitebox access to any detector. While better results could be achieved with additional prompt fine-tuning and human editing, we stand behind our results as an important proof-of-concept showing the weaknesses of these AI text detectors.

However, we acknowledge that the biggest limitation of our work was that we did not fully `break` GPTZero. The decision boundary we chose for our experiments—less than 50\% ``AI Probability'' for a negative classification—was a more lenient threshold than the \textit{AI}, \textit{Mixed}, and \textit{Human} labels returned by the GPTZero classifier. It is likely that a user (e.g. a teacher checking their students' work) will focus primarily on the labels and use the AI probability score only as a supplementary metric. Under the stricter regime of requiring a \textit{Human} label for a negative classification, our current attacks do not succeed, with only one of 400 AI-generated texts being classified as \textit{Human}. Nevertheless, we do not believe that this is definitive evidence that GPTZero is robust in all cases. Our paraphrasing attacks show that paraphrasing \textit{can} be used to lower the ``AI Probability'' score of AI-generated texts, and so we believe that further prompt engineering will be successful at deceiving the detector even with the stricter classification threshold. In particular, we identify balancing sentence perplexity with text simplicity and readability as one of the primary challenges for a paraphrasing attack to be successful.

The second limitation of our work relates to text quality. While the high USE cosine similarity scores indicate that the paraphrased texts are semantically similar, the similarity metric does not necessary indicate the linguistic quality of the text. Further work is needed to develop metrics for writing style, register, and vocabulary. For example, although the \textit{College Student} paraphrasing attack has a 0.83 cosine similarity with the original ChatGPT text, manual inspection of these paraphrases reveal that the writing style is distinct from the human and GPT essays, since this paraphrasing attack uses an informal register and colloquial vocabulary that is easy to identify. Thus, while this paraphrasing attack deceives the AI text detectors, it does so by noticeably altering important stylistic characteristics of the text that could be easily identified by a human (e.g. the teacher reading their students' work). Furthermore, our watermarked GPT Neo generations suffered from low text quality and the widespread presence of non-Latin characters, an example of which can be see in Appendix \ref{detector-descriptions}. However, we view this as a limitation of watermarking itself, since the need for whitebox access to the LLM limits which models can be used for generation (i.e. newer, more powerful models cannot currently be watermarked).

\subsection{The ethics of AI text detection}

Our intention in pursuing this research was to evaluate AI text detectors in the particular context of students using LLMs to write academic assignments, and teachers using off-the-shelf AI text detectors to analyze student work. As we suspected, the vulnerabilities of the detectors we evaluated raise significant questions about the ethics of relying on these detectors to make determinations about academic dishonesty. As we note in Appendix \ref{detector-descriptions} and section \ref{detector-claims}, the detectors we use make claims about accuracies and error rates that are not substantiated by our results. When the potential for harm is this high, we caution against the use of these detectors as the \textit{primary} indicator of whether a student should be accused of academic dishonesty, especially since human inspection of the paraphrased texts is in many cases sufficient to raise suspicions that the text was LLM-generated.

\bibliographystyle{plain}
\bibliography{main}


\appendix

\section{Details of AI text detectors}\label{detector-descriptions}

\subsubsection{Kirchenbauer et al. watermarking}

As described in section \ref{watermarking}, the Kirchenbauer et al. watermarking algorithm partitions tokens in a vocabulary into a smaller \textit{green list} and larger \textit{red list} according to the desired proportion of green list words, $\gamma$. At generation time, the algorithm preferentially samples from the green list, thereby embedding the watermark in the text. At detection time, a p-value is generated based on the proportion of green listed tokens in the text, and the detector classifies a text as AI-generated if the p-value is below a specified threshold. To replicate the setup of Kirchenbauer et al. \cite{kirchenbauer_watermark_2023, kirchenbauer_reliability_2023}, we use the same parameter values, including $\gamma = 0.25$.

\begin{figure}
    \centering
    \includegraphics[width=10cm]{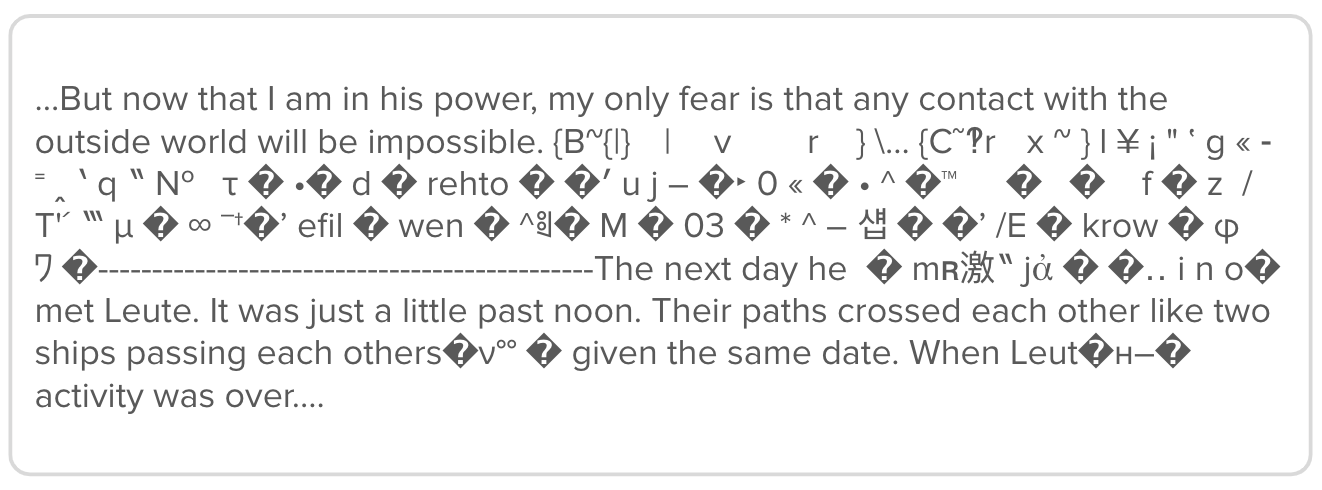}
    \caption{An excerpt from a watermarked GPT Neo text that includes multiple textual artifacts.}
    \label{fig:watermarking-artifacts}
\end{figure}

Although we were able to replicate the Kirchenbauer watermarking regime for our experiments (using the code provided in the Kirchenbauer papers), we were not always satisfied with the quality of the generated text. Figure \ref{fig:watermarking-artifacts} provides one example of watermarked text that includes multiple non-Latin alphabet characters, a phenomenon that occurred in many generated texts. While we are cognizant that our use of GPT Neo as a foundation model for watermarking has its inherent limitations, we note that watermarking requires whitebox access to the generating model's logits in order to implant the watermark. Thus, the only models that can be watermarked are those that allow whitebox access, limiting current watermarking efforts to older and less powerful models. We believe that this limitation and these generation artifacts merit further investigation.

\subsubsection{ZeroGPT}

ZeroGPT is a free online AI text detector that advertises itself as ``the most Advanced and Reliable Chat GPT, GPT4 \& AI Content Detector,'' The website allows the user to paste in or upload a text of up to 15,000 characters, and returns both an ``AI GPT'' percentage, and a classification ranging from ``Your text is Human written'' to ``Your text is AI/GPT Generated,'' The website claims that ``After analyzing more than 10M articles and text, some generated by AI and others written by humans, we developed ZeroGPT's algorithm with an accuracy rate of text detection up to 98\%,''

\begin{figure}
    \centering
    \includegraphics[width=8cm]{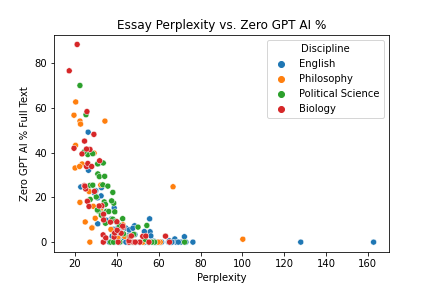}
    \caption{The scatter plot shows the strong ($r = -0.60$) negative relationship between a human text's perplexity and its ZeroGPT ``AI \%'' rating.}
    \label{fig:perplexity-scatter}
\end{figure}

We suspect that ZeroGPT relies largely on the perplexity of the candidate text in order to make a classification. Figure \ref{fig:perplexity-scatter} shows a negative relationship between essay perplexity and ZeroGPT ``AI \%'' scores across all disciplines of human essays. The correlation coefficient for this relationship is $r=-0.60$.
Thus, if a text's perplexity is closer to human-level average perplexity, then the detector will misclassify an AI-generated text as human-written. This provides a clear explanation for why all paraphrasing attacks achieved a high attack success rate against ZeroGPT. 

\subsubsection{GPTZero}

GPTZero is another online AI text detector, with pricing tiers (including a free option) that afford different daily or monthly text limits. The detector allows the pasting or uploading of a text, and returns a predicted ``AI Probability'' as well as an overall classification from one of the following three labels: \textit{AI}, \textit{Mixed}, \textit{Human}. The ``AI Probability'' score is defined as the ``probability this text was entirely written by AI,'' Initial exploration of this detector revealed difficulty in fooling the detector into classifying an AI-generated text as ``Human'', so for this project we focused on the intermediate goal of bringing the ``AI Probability'' score below 50\%. GPTZero emphasizes that its AI text detection technology is ``trusted'' by organizations such as the learning management system Canvas, and universities including UC Berkeley and NYU. In October 2023, GPTZero announced\footnote{\href{https://gptzero.me/news/gptzero-partners-with-american-federation-of-teachers-aft-to-support-responsible-ai-adoption-in-classrooms}{https://gptzero.me/news/gptzero-partners-with-american-federation-of-teachers-aft-to-support-responsible-ai-adoption-in-classrooms}} a partnership with the American Federation of Teachers (AFT) to bring its AI tools to classrooms, including unlimited AI detection access for 15,000 teachers. Because of the significant rate limiting of GPTZero, we use a random subset of 20 essays from each discipline (n = 80) to test this detector. We also bought the lowest tier paid plan for \$15 per month to increase the total number of characters we could use with the detector. In total we spent \$30 across two months.

\section{Text examples}\label{text-examples}

Table \ref{essay-table} shows the original human, the original GPT, and the \textit{perplexity}, \textit{college student} and \textit{recursive} paraphrases of an English essay titled ``Historical Places, Violent Spaces: A discussion on violence, personhood, and space in Virginia Woolf's Between the Acts''.
 
\begin{table}
\caption{Human and GPT versions of an English essay}
  \label{essay-table}
  \centering
  \small
\begin{tabular}{lp{10cm}}\toprule

    Text Type & Text \\\midrule

    Human Essay & Virginia Woolf's last novel Between the Acts is, more than anything else, a comment on contemporary society and how individuals function within that structure. She uses the house of the Olivers and the small-town pageant to provide a microcosm of the pressures and issues that existed in pre-World War II England. Both the characters and the house, the main focuses of the novel, have their own history of class and culture and attempt to distinguish themselves among the influences and voices of the past. Each character has a way of negotiating the pressures of the past; what the narration becomes is a portrait of people's acceptance of and resistance to their own individual chains of destiny—specifically regarding cycles of violence in history. In Between the Acts, the concept of space acts as a vehicle through which a study of character formation and the difference between individual assertions of place and the ways that history, culture, and one's own past limit freedoms of self-assertion. \\\midrule
    Original GPT Generation & Virginia Woolf's novel "Between the Acts" is a literary tapestry woven with threads of history, personal reflections, and an exploration of the violence that exists within the boundaries of both public and private spaces. Published in 1941, the novel captures the essence of an England on the cusp of transformation, and it delves into the ways in which history, violence, and the concept of personhood intersect and collide within the context of space. This essay seeks to dissect the intricate interplay of these themes in Woolf's work, illuminating the profound impact they have on the characters and the narrative. \\\midrule
    \textit{Perplexity} Paraphrase & Virginia Woolf's literary creation, "Between the Acts," intricately weaves together strands of history, personal introspection, and an examination of the inherent violence that permeates both public and private domains. Published in 1941, the novel encapsulates the essence of an England teetering on the precipice of profound transformation, delving deep into the ways in which history, violence, and the concept of personhood intermingle and collide within the expansive realm of space. This essay endeavors to deconstruct the intricate interplay of these overarching themes in Woolf's opus, shedding light on the profound ramifications they exert upon the denizens of the narrative tapestry. \\\midrule
    \textit{College Student} Paraphrase & Virginia Woolf's book "Between the Acts" is like this literary tapestry that's all about history, personal reflections, and diving into the violence lurking in both public and private spaces. Published back in 1941, this novel really captures the vibe of England on the brink of change. It digs into how history, violence, and the whole idea of personhood mix and clash within the space scene. This essay is all about breaking down how these themes play out in Woolf's work and how they shake up the characters and the story. \\\midrule
    \textit{Recursive} Paraphrase & Virginia Woolf's work, "Between the Acts," intricately blends history, personal reflection, and an exploration of inherent violence in both public and private realms. Published in 1941, the novel captures England on the brink of significant change, delving into how history, violence, and individual identity intersect within the vast expanse of space. This essay aims to unravel the complex interplay of these themes in Woolf's masterpiece, shedding light on their profound impact on the characters in the narrative. \\
    \bottomrule

\end{tabular}
\end{table}

\section{Semantic similarity scores}\label{semantic-similarity}

Table \ref{use-table} shows the cosine similarity scores for the Universal Sentence Encoder embeddings of different combinations of human, GPT and paraphrased texts.

\begin{table}
\caption{Universal Sentence Encoder cosine similarities}
  \label{use-table}
  \centering
  \small
\begin{tabular}{lc}\toprule

    Texts & Avg. Cosine Similarity \\\midrule

    Human Text vs. Human Text & 0.28 \\
    Human Text vs. Original GPT Generation & 0.63 \\
    Original GPT Generation vs. \textit{Perplexity} Paraphrasing & 0.88 \\
    Original GPT Generation vs. \textit{College Student} Paraphrasing & 0.83 \\
    Original GPT Generation vs. \textit{Recursive} Paraphrasing & 0.95 \\
    \bottomrule

\end{tabular}
\end{table}

\end{document}